\def\year{2019}\relax
\documentclass[letterpaper]{article} 
\usepackage{aaai19}  
\usepackage{times}  
\usepackage{helvet} 
\usepackage{courier}  
\usepackage[hyphens]{url}  
\usepackage{graphicx} 
\usepackage{graphicx}  
\begin{document}
%
\title{Get IT Scored Using AutoSAS - An Automated System for Scoring Short Answers}


\author{
Yaman Kumar,\textsuperscript{\rm 1}
Swati Aggarwal,\textsuperscript{\rm 2}
Debanjan Mahata,\textsuperscript{\rm 3}
Rajiv Ratn Shah,\textsuperscript{\rm 4} \\
\bf \Large Ponnurangam Kumaraguru,\textsuperscript{\rm 4}
Roger Zimmermann\textsuperscript{\rm 5}\\
\textsuperscript{\rm 1}Adobe, India,
\textsuperscript{\rm 2}NSIT-Delhi, India,
\textsuperscript{\rm 3}Bloomberg, USA \\
\textsuperscript{\rm 4}IIIT-Delhi, India,
\textsuperscript{\rm 5}National University of Singapore, Singapore\\
ykumar@adobe.com, swati@nsit.ac.in, dmahata@bloomberg.net. 
rajivratn@iiitd.ac.in, \\ pk@iiitd.ac.in, rogerz@comp.nus.edu.sg
}

\maketitle
\begin{abstract}
In the era of MOOCs, online exams are taken by millions of candidates, where scoring short answers is an integral part. It becomes intractable to evaluate them by human graders. Thus, a generic automated system capable of grading these responses should be designed and deployed. In this paper, we present a fast, scalable, and accurate approach towards automated \textit{Short Answer Scoring} (SAS). We propose and explain the design and development of a system for SAS, namely AutoSAS. Given a question along with its graded samples, AutoSAS can learn to grade that prompt successfully.  This paper further lays down the features such as \textit{lexical diversity}, \textit{Word2Vec}, \textit{prompt}, and \textit{content overlap} that plays a pivotal role in building our proposed model. We also present a methodology for indicating the factors responsible for scoring an answer. The trained model is evaluated on an extensively used public dataset, namely \textit{Automated Student Assessment Prize Short Answer Scoring} (ASAP-SAS). AutoSAS shows state-of-the-art performance and achieves better results by over 8\%  in some of the question prompts as measured by Quadratic Weighted Kappa (QWK), showing performance comparable to humans.

\end{abstract}

\section{Introduction}
Essays and other types of writing practices have been extensively used for evaluation purposes. Graduate Record Examination (GRE), Scholastic Aptitude Test (SAT), Senior School Examinations such as Zhongkao in China and All India Senior School Certificate Examination (AISSCE) in India are just some of the many examples. The stakes for getting high grades in the essays and hence in  these exams are tremendous for \textit{pupils}, \textit{teachers} and \textit{schools} alike. The essays and short answers written by the students in the exams determine their future colleges and hence have a career wide impact. 

Under the \emph{No Child Left Behind Regulations}, U.S. States have been asked to use uniform and regulated test scores for evaluation of teachers for determining their salaries and tenures \cite{higgins_2014}. This underlines the importance of getting good scores in these writing practices. A school's reputation is often determined by the SAT scores of its graduating students, which in turn is impacted by how well they have been taught to write their essays and short answers \cite{dale2002estimating}.

\subsection{Motivation} \textit{Automated Essay Scoring} (AES) and \textit{Short Answer Scoring} (SAS) systems such as the one presented in this work (AutoSAS), provides economic advantages to testing companies and state-wide corporations. These systems reduce the economic and time burden of getting each response checked by human graders close to zero, bringing down the cost and effort significantly. 

It has been noted that about 30\% of a teacher's time is spent in evaluating students that subsequently translates to close to 4.02 Billion US Dollars per year coming from the taxpayers \cite{mason2002automated}. To eliminate this, it is necessary to design an automated system that a teacher can trust, and can use to mark essays and short text responses. In addition, one often hears about biases in marking students based on region, religion, and ethnicity. AES systems can possibly aid in uprooting any such biases from the education system. 

Currently, AES systems have been successfully deployed by Educational Testing Service\footnote{https://www.ets.org/}, where GRE essays are graded by a human grader as well as an AES system \cite{burstein1998computer}. A second human grader is required only if there is a non-negligible difference between the two grades. AES systems form a major use case for Massive Open Online Classes (MOOCs) where economies of scale are required. As MOOCs advent towards offering courses in subjects such as literature and humanities, a range of assessment techniques such as AES will come in handy. 

Systems similar to AutoSAS can be deployed in other agencies where it can not only be used for reducing the economic cost related to grading, but also in providing a scalable system for uniform grading in a time-bound manner. Students can also benefit from use of these systems if they can verify and check their work before actually submitting it for final reviews.

  In this work, we primarily focus on utilizing natural language processing (NLP) for the task of Short Answer Scoring (SAS), which involves automated scoring of short answers provided for a given prompt that presents questions from a fixed set of subjects. This problem is typically formulated as a supervised learning problem where samples are graded on an ordinal scale (say, 1-10). AES as a NLP problem has been studied extensively ~\cite{balfour2013assessing,xi2010automated,valenti2003overview,yang2002review}, 
with a variety of methods being deployed to perform the task of grading. We attempt to solve it as a supervised regression problem and develop a system, namely Automated Short Answer Scoring (AutoSAS) on top of a popular publicly available dataset.

The quality of text from the perspective of scoring short answers is dependent on many factors, some of them being \textit{content}, \textit{grammar}, \textit{vocabulary}, \textit{flow}, \textit{coherence of ideas} and \textit{relevance to the topic}. We propose a novel model utilizing many of these elements to grade short answers. A thorough overview of all the features used in building AutoSAS is presented. Using these features, AutoSAS performs better than the current state-of-the-art models for grading students' short answers. For some of the prompts, the improvements are more than 8\%. 


AutoSAS provides a listing of all the features that are important for the grading of a particular response in a ranked manner. It also presents a listing of the features which contributed to the score of a particular candidate. Through this raw feedback, the students can assess their weak areas. This may serve as an invaluable feedback for the students. Using the feedback, they can improve their writing before submitting the final document. This can also be used as an alternative as well as to augment the feedback that teachers provide while they grade student responses.

\subsection{Contributions} Towards the objective of scoring short answers we make the following contributions in the work presented in this paper:
\begin{enumerate}
\item We present a supervised model for automatically scoring short answers that shows state-of-the-art performance with improvements of more than 8\% in certain sets of question prompts as measured by Quadratic Weighted Kappa (QWK).
\item We augment existing features used by previous works with new set of features along with their ablation study.
\item AutoSAS as a system can not only score short answers but can also possibly find its usage in providing feedback to its users about their response, giving a detailed overview of what went behind its decision making process.
\end{enumerate}

Next, we briefly present some previous research and systems that are relevant to the scope of our work in this paper.
\section{Related Work}
\label{research_background}

Different authors and organizations have ventured into building AES systems extending the Project Essay Grade (PEG) \cite{page1994computer}, which is one of the first systems developed. AES is ordinarily considered as a regression or a classification problem. In regression based analysis, essay score is considered to be a dependent variable and depends on values of features of an essay. These features are then used to learn a regression equation which is further used for grading essays. 

In classification based approaches, essays are segregated into different classes according to their scores. These classes then form the basis for segregating and scoring the future essays. Different techniques and models have been developed on various datasets. \textit{E-Rater}, a system designed by Educational Testing Services (ETS) \cite{attali2006automated}, utilizes stepwise regression analysis on diverse linguistic features. It is used in popular exams like GRE and Test of English as a Foreign Language (TOEFL).

Apart from standard features such as \textit{TF-IDF}, \textit{word frequency analysis}, many have attempted to use more diverse and heterogeneous set of features for this task such as \textit{lexical chain} \cite{somasundaran2014lexical}, \textit{students' demographic information}, \textit{reading comprehension}, \textit{vocabulary knowledge}, \textit{writing apprehension}, among others for scoring essays \cite{crossley2015pssst}.

There have been many previous attempts to automate the scoring of short answers as well as essays. The methods utilized have ranged from regression to classification based supervised learning. In spite of the breadth, the previous works considered a very restrictive set of features \cite{chen2010unsupervised} which were often hand-picked and were restrictive to the domain they were applied to \cite{ramachandran2015identifying,bachman2002reliable,riordan2017investigating}, compromising the reliability and accuracy of predictions substantially. 

We propose a supervised model, namely AutoSAS, which is developed to grade short answers. This model shows a significant improvement over the current state of the art technologies in the accuracy of predictions and scalability over disparate domains. In summary, this paper presents a simple to use, fast and reliable approach to grade short answers that can be easily used in a classroom setting. 

\section{Task Overview}
\label{task_overview}
In this section, we present the details of the dataset used, a comparison of AES and SAS, and the reasons behind SAS being more difficult in nature. Then, we detail the features used by AutoSAS for grading short responses.

\begin{table}[htbp]
\label{dataset_overview}
\begin{tabular}{|l|l|}
\hline
\multicolumn{1}{|c|}{\textbf{Topic}} & \multicolumn{1}{c|}{\textbf{Question}} \\ \hline
Science & \begin{tabular}[c]{@{}l@{}}Replicate an experiment based on the details of\\ another experiment\end{tabular} \\ \hline
Arts & \begin{tabular}[c]{@{}l@{}}Similarity between Pandas, Koalas and \\ differences wrt. Python\end{tabular} \\ \hline
Biology & Describe protein synthesis \\ \hline
English & Describe a character Mr. Leonard \\ \hline
\end{tabular}
\caption{Sample questions from the dataset.}
\end{table}

\subsection{Dataset} We conduct the experiments on a public dataset released by Automated Student Assessment Prize (ASAP) competition hosted in Kaggle\footnote{\url{https://www.kaggle.com/c/asap-sas/}}, and was sponsored by Hewlett Foundation. This is the largest publicly available dataset, consisting of student responses for a total of 10 different questions and more than 16000 responses. This dataset has also been popularly used by researchers who have reported works similar to us \cite{ramachandran2015identifying,riordan2017investigating}. 

The responses were written by high school students and then manually graded and double scored (on a scale of 0-3) by the ASAP graders. The questions covered a range of topics from Science to Language and Arts. A brief overview of the questionnaire is presented in Table 1. The questions belonged to a variety of topics, with their response length ranging from 1 word to 300 words, with an average of 50 words. Information provided in the responses ranged from the question itself (verbatim in some cases) to the author's preformed knowledge. Due to the realistic nature (non-lab environment) and diversity of the dataset, it is ideal for our analysis.


\subsection{Automatic Essay vs Short Answer Scoring:}
Although AES and SAS as NLP tasks have a lot in common, yet SAS is significantly different from AES in following ways:\begin{itemize}

\item \textbf{Length of Response} : Short answers are typically shorter in length than essays. This means, essentially, that the author is constrained to present his ideas in a shorter response, and has scope to present fewer ideas. This is a challenge for an automated system, as there are lesser number of tokens (words, phrases \emph{etc.}) that are related to the domain about which the writer is writing \cite{ramachandran2015identifying}.

\item \textbf{Genre} : Essays emphasize on \textit{narration} and \textit{imagination}, which is not possible in short answers since they are required to be precise and to the point. The model short answers should cover all the major points respecting the space provided. For a concrete example, an intrigued reader can access the marking scheme of the questions asked in the datasets from the data set description.


\end{itemize}
Next, we describe the various groups of features that were extracted from the responses and forms the basis of our trained model and further analysis. Wherever applicable we point to the previous studies that has used any of them.
\subsection{Features\label{features}}

\begin{itemize}
\item \textbf{Word2Vec and Doc2Vec based Features:}
\label{featuresExtracted_word2vec}
Word2Vec \cite{mikolov2013distributed} and Doc2Vec \cite{le2014distributed} are useful techniques that capture semantic relationships between words and documents from the different contexts in which they occur. We used pre-trained Word2Vec and Doc2Vec models trained on Google News corpus and Wikipedia dump, respectively. These corpuses are used so as to model the generic nature of the question prompts. The questions in the dataset ranges from Science to Arts and Literature. Sentences from such a variety of topics can only be covered by a corpus such as that of news reported from different domains and encyclopedic entries on a vast range of topics. Word2Vec although provides high quality word vectors but averaging them makes them lose their order information. Thus Doc2Vec was also used over the full short answer.


\item \textbf{Part of Speech (POS) Tagging:}
POS tags based n-grams capture context very well. Each response word was tagged with its corresponding part-of-speech (eg., Verb, Noun, Preposition). 
To take care of the context information, POS-tagged bi-grams, tri-grams and tetra-grams were extracted from the prompt. 
To avoid the trivial n-grams, based on the training set, a list of significant bi-grams, tri-grams and tetra-grams was constructed. For constructing such a list, we used the responses that were graded high by the human graders. For a particular response, the n-grams which are present in this set were considered and all others were ignored. The final selection of the n-gram set was based on considering only those that have greater than a particular incidence count. The threshold was determined by running the model on validation data, keeping other features constant.


\item \textbf{Weighted Keywords:}
\label{featuresExtracted_Google}
Certain prompts demand a set of domain-specific keywords to be present in them. 
For getting a list of keywords, we took following steps. Firstly, a set of domain-specific words were identified from the set of answers to questions related to a specific subject. Then using Google API, the top 20 articles related to that word were extracted. Each page was subsequently scraped and the term frequencies of keywords were stored. Then, tf-idf values were calculated for each word, thus getting a list of keywords based on their tf-idf importance.

\item \textbf{Prompt Overlap:}
Short answers derive some of their information from the question itself. Therefore, an overlap between the prompt and the response serves as an important grading metric for any grader. 
As an example, in reading comprehension based prompts, the answers derive their context and content from the comprehension and the question itself. This content can be in the form of taking information of subject, verb, argument, time, from the question itself and then using this information to answer it. For example, one of the questions given in the dataset,  ``Based on Rose's conversations with two other characters, describe her.''  This question requires an answer that includes context from the text of the question. Thus an overlap between the answer and a question is expected and necessary.

\item \textbf{Lexical Overlap:}
In many of the prompts, even after taking prompt overlap, many words that can be found in the questions such as reading comprehensions were not captured. These words can only be extracted from the comprehension itself, neither Google keywords, nor a simple overlap with question prompt would yield something significant. Thus, AutoSAS takes into consideration different types of lexical overlaps between the sentences present in the short answers: \textit{Noun Overlap}, \textit{Argument overlap}, and \textit{Content Overlap}. 

\textit{Noun Overlap} is a measure of the frequency of overlap of nouns across two sentences. \textit{Argument Overlap} is measured by the overlapping intersection of arguments among sentences. We extract argument from a sentence using a list of hand-crafted heuristics. \textit{Content overlap} measures the amount of overlap of content words across sentences. Wherever applicable, this set was further extended to include the words' synonyms using WordNet\footnote{https://wordnet.princeton.edu/}. All the overlaps are calculated w.r.t the reading comprehension text. For example, in case of noun overlaps if there are 5 nouns in the comprehension text and the answer has 3 nouns the overlap is calculated to be $3/5$. These scores are further normalized across individual prompt sets.

Previous studies show that lexical overlap significantly aids in text analysis. \cite{rashotte1985repeated,ferris1994lexical,douglas1981exploratory}. 
For instance, response no. 19953 in the dataset states, ``Paul finds out that Mr. Leonard was a track star but he could not read. `No school wanted a runner who couldn't read.' '' Thus the \textit{Nouns}- Mr. Leonard, school runner, Paul match with what is given in the comprehension. The overlap between the \textit{Argument}, ``No school wanted a runner....'' matches with the prompt. This is one of the reasons of this response's high grades (it scored a 2).

\item \textbf{Word Frequency, Difficulty and Diversity:}
Word frequency and diversity indicate a student's command over language. Although, many earlier works \cite{nation1996vocabprofile}, use this technique but they look at the frequency of top \emph{k} words only. We find this approach non-holistic. This is so since the top words (frequency wise) can be more quickly accessed by a writer and thus are consequently easier to decode by a reader \cite{perfetti1985reading,rayner1989}, whereas, as indicated by many studies \cite{frase1998computer,reppen1995variation,reid1990responding} writers using lesser frequent words are, generally, more proficient than others. Thus with this in mind, Webster Dictionary\footnote{\url{https://www.https://www.merriam-webster.com/}} was divided into 20 different levels of words with varying difficulties \cite{breland1994college}. Then each word used by the candidate was mapped to a difficulty level using this dictionary. The frequencies of words for each of the difficulty levels were noted as features of that response.

In addition, the number of unique words that appear in each response and \textit{Type Token Ratio} (TTR) \cite{templin1957certain} serve as another set of features. TTR is a value which ranges from 0 to 1 and is indicative of the lexical diversity of a prose. The writer with a larger vocabulary is generally more proficient and hence is better graded than a writer using limited vocabulary \cite{engber1995relationship,reppen1995variation}. 

\item \textbf{Statistics of Sentence and Word Length:}
In general, sentence length indicate the complexity of the sentence, with longer sentences requiring greater use of working memory and hence being more difficult to understand. Thus word and sentence length can be used to indicate the sophistication of a writer \cite{hiebert2011beyond}. Several research projects have shown that higher-rated essays, in general, contain more words \cite{carlson1985relationship,ferris1994lexical,reid1990responding} 
and generally use longer words \cite{frase1998computer,reppen1995variation}. Thus sentence length, word length, average sentence and word length were noted down for each response and used as features.

\item \textbf{Logical Operators based Features:}
Logical operators are directly related to the number, density and abstractness of ideas of a student. These then translate to the quality of arguments in a text \cite{fayol1997acquiring}.  Some other studies have also used them, for example, Coh-Metrix \cite{crossley2012predicting}. AutoSAS uses \textit{and, or, not, if-else, if-then, unless, whether, although, but} and their various other combinations as logical operators for grading purposes. As an example, response with Id 231, states, ``\textit{If} I used different amounts of water \textit{when} washing the samples, one may not be as thoroughly washed as another \textit{which} could mess up the results.'' The logical operators combination \textit{if}, \textit{when} and \textit{which} indicate the logical complexity of the sentence.

\item \textbf{Temporal Features:}
Temporal features such as tense and aspect words help the grader in forming a timeline of events thus enhancing the validity and pithiness of the arguments of a student. While tense helps us in formation of a sequence, aspect represents the dynamics of the events with respect to time \cite{klein2013time}. It has been argued \cite{mccarthy2007coh} that repeating tenses and aspects in the text create more cohesion in the arguments presented in a response, hence improving its quality. Some other studies \cite{crossley2012predicting} have also used aspect and tense based features to extract temporal features. Thus various tense and aspect words were identified in a response and then, were associated with events. 

For instance, response with Id 8078, responded, ``I \textit{believe} in this article 'invasive' \textit{means} hidden/unchecked or \textit{referred} to pythons MaccInnes \textit{uses} this word when he \textit{states} ,`` I \textit{think} that invasive \textit{is passing} judgement" he \textit{used} this because he in happy that pythons \textit{are going} birth....'' Due to wrong usage of tenses, among other reasons, it scored a 1.

\end{itemize}

\begin{figure}[htbp]
  \includegraphics[width=1.0\columnwidth]{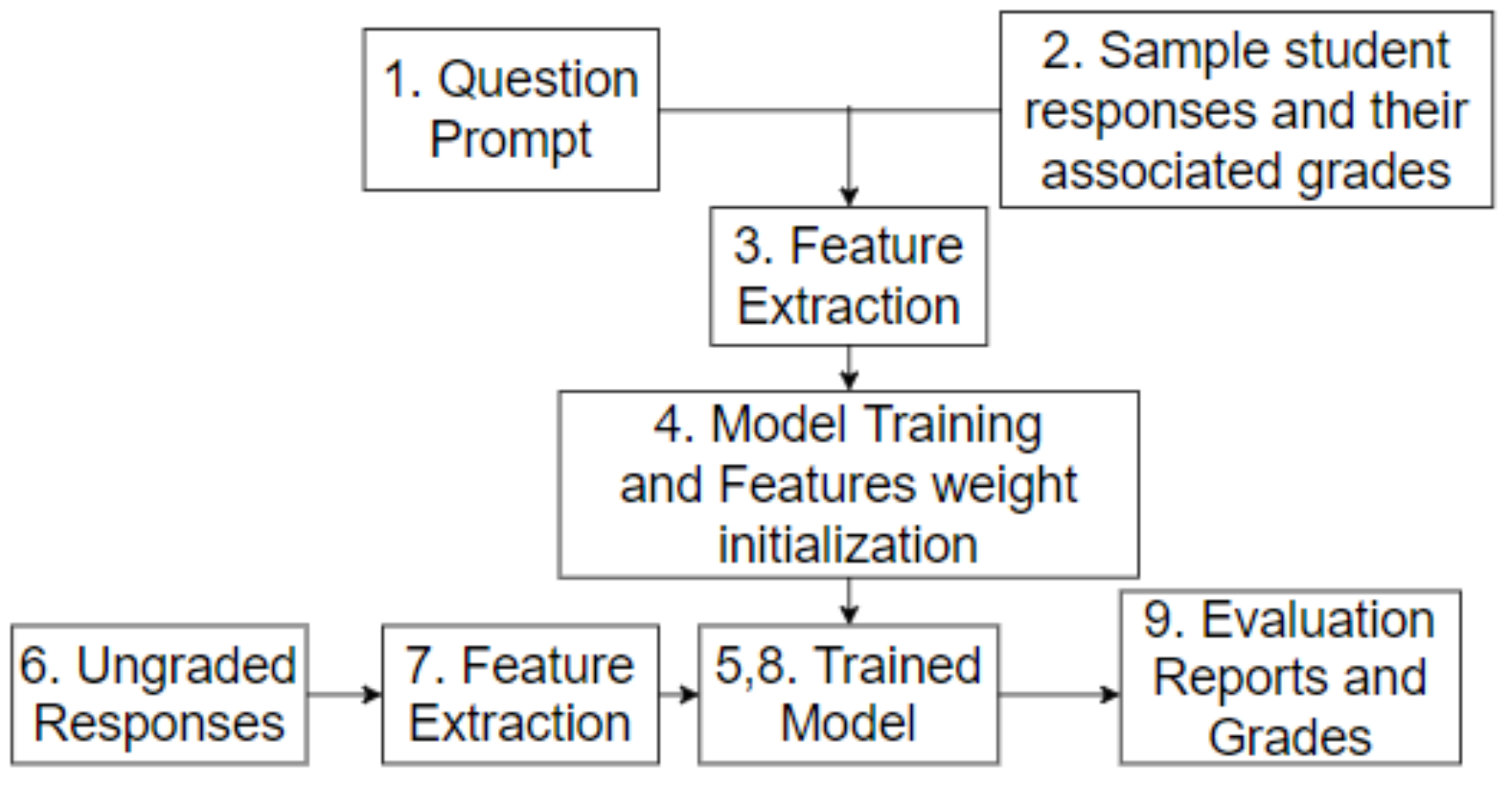}
  \caption{Pipeline for scoring short answers using AutoSAS.}
  \label{fig:training}
\end{figure}

\section{Experiments}
\label{experiments}

In this section, we provide a detailed description of the supervised training process of our model. The complete pipeline of the proposed model, AutoSAS is shown in Figure \ref{fig:training}. For training the model, it is required that the users (\emph{i.e.} teachers, schools, interviewers, \emph{etc.}) grade some of the responses, and provide them as an input to AutoSAS along with their respective question prompts. The question prompt is required for extracting features, as explained in the previous section. Different sets of question prompts will have different number of features. The sample responses given to train the model will help to finalize the weights of these features, which will then be used for grading of the ungraded responses. Once the grading has been done successfully, it also produces feedback for a given response.

\subsection{Augmentation of Training Set using Jumbled Content}
\label{dataset_augmentation}
For our experiments, we also augmented the training set using jumbled content. Two types of jumbled content were included along with the normal responses in order to train AutoSAS for a particular prompt. This was done in order to avoid grading those responses highly, that were written well, but had content irrelevant to the question asked. 

We included 10 highly rated responses from prompts other than the one for which the model is being trained for and gave them the lowest grade possible. This made sure to penalize the irrelevant responses. Another source for jumbled content was from the answers of the same prompt on which AutoSAS was being trained. After an initial training for that particular prompt, some of the these otherwise highly graded responses were taken, jumbled up and then included with the training samples assigning them lowest possible grades. This was done to avoid grading those responses highly, which included a soup of gibberish keywords related to the question while not having context, connecting information \cite{perelman2014state}.

\subsection{Training the Model} Firstly, all the responses for a particular prompt are prepared. The responses are obtained either from the dataset corresponding to a particular prompt or are obtained via dataset augmentation. Then, each response is checked for spelling mistakes as it is mentioned in most of the grading rubrics grammar, penmanship and spellings were important for the clarity of responses but were not important for scoring. Words involving scientific names such as chemical compounds, proper nouns and other tokens which are not found in the dictionary but are employed in the responses are taken care of appropriately during spell correction. The set of features as described previously were subsequently extracted from the responses and the questions. Responses from different prompts were saved separately and subsequently loaded in different dataframes.

In order to train AutoSAS for a particular question, all the features and grades of that question are loaded in a dataframe, which is then used for \textit{regression analysis}. For splitting the total data into train, validation and test data, a ratio of 70:10:20 was used. It is to be noted that none of the jumbled responses were included in the test data. The test set solely consisted of the original data. Other type of data such as jumbled responses were included solely for training purposes. Testing and training data was stratified so as to get an equalized distribution of samples across all grades. 

\begin{center}
\begin{figure}[htbp]
\centering
\includegraphics[width=1.0\columnwidth]{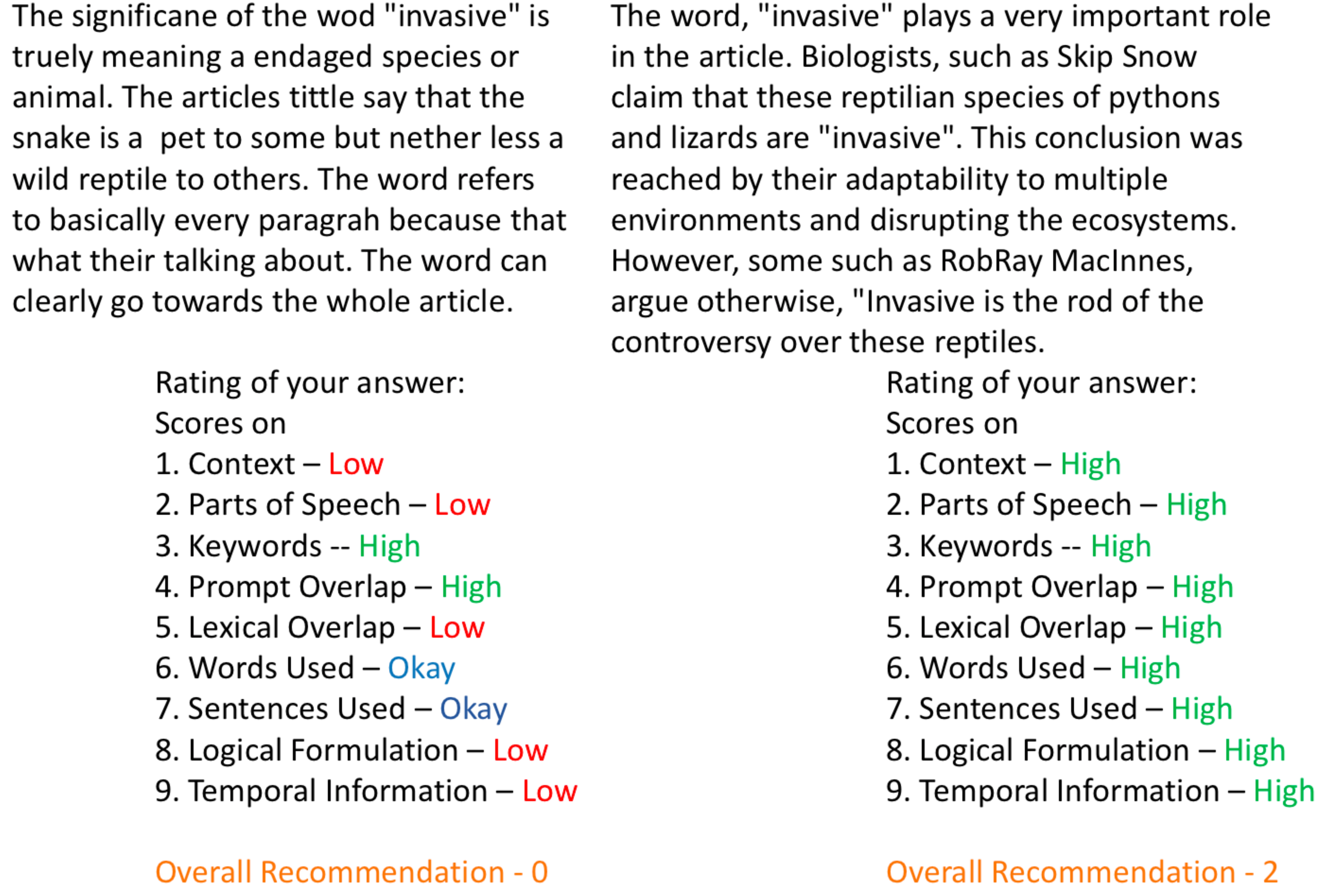}
\caption{An example of feedback produced by AutoSAS.}
\label{fig:fedback_0}
\end{figure}
\end{center}

A \textit{Random Forest} model was trained on all the features. Random forest model has been used chiefly because of two reasons. Firstly, it performs well over the multitude of features extracted from the responses (as presented in the results). Secondly, in order to know the importance of each feature set used in the analysis, which is not possible with popularly used neural network based approaches (as presented in Ablation study). An example of feedback produced by AutoSAS is given in Figure \ref{fig:fedback_0}. With multi layer neural networks (RNN, CNN), one can feed the network a listing of all the features but cannot easily expect the network to tell the performance of each feature for a particular response  \cite{leray1999feature,montavon2017methods}. 


In the case of \textit{Random Forest} model, we get the performance of all the features on a particular response by making use of the package TreeInterpreter\footnote{\url{https://github.com/andosa/treeinterpreter}}. It gives a list of contributions of each feature in getting the grade for a particular response. For a task of such nature as grading is, students expect the teacher to give atleast a crude indication of the reason behind the scores assigned to them. This feedback might be useful for a student's improvement. This makes \textit{interpretability} of the model an important task. Though we do not explore an exhaustive mechanism for comments and review part of the grading process, however, a crude indication of the scores assigned is presented. 

As a part of our future work, we would like to explore the proposition of giving a more exhaustive explanation for the grades assigned, similar to some other recent work \cite{chen2018learning} dedicated to this goal. With a \textit{Random Forest} model, AutoSAS presents a listing of all the features computed for a particular response along with the importance of each such feature in its grading process. Thus, the author of the response has the opportunity to know what are his/her weak areas, and can focus on them. The results were computed using the process mentioned, and are presented in the next section along with a comparison of the current state-of-the-art models.

\begin{table*}[htbp]
\centering

\label{comparison_table_1}
\begin{tabular}{|l|l|l|l|l|l|l|l|l|l|l|l|}
\hline
\textbf{Approach}           & \textbf{Set 1} & \textbf{Set 2} & \textbf{Set 3} & \textbf{Set 4} & \textbf{Set 5} & \textbf{Set 6} & \textbf{Set 7} & \textbf{Set 8} & \textbf{Set 9} & \textbf{Set 10} & \textbf{Mean}  \\ \hline
\textbf{AutoSAS}            & \textbf{0.872} & \textbf{0.824} & \textbf{0.745} & \textbf{0.743} & \textbf{0.845} & 0.858          & \textbf{0.725} & 0.624          & \textbf{0.843} & \textbf{0.832}  & \textbf{0.791} \\ \hline
\textbf{Ramachandran et al.} & 0.86           & 0.78           & 0.66           & 0.70           & \textbf{0.84}  & \textbf{0.88}  & 0.66           & \textbf{0.63}  & \textbf{0.84}  & 0.79            & 0.78           \\ \hline
\textbf{Riordan et al.}     & 0.795          & 0.718          & 0.684          & 0.700          & 0.830          & 0.790          & 0.648          & 0.554          & 0.777          & 0.735           & 0.723          \\ \hline
\end{tabular}
\caption{Comparison of performance of models on the dataset, ASAP-SAS. The data presented is QWK scores for each of the ten prompts in the dataset.}
\end{table*}

\section{Results}
\label{results}

\subsection{Evaluation Metrics}
We use Quadratic Weighted Kappa (QWK) \cite{brenner1996dependence} as the evaluation metric for finding the agreement between the grades predicted by AutoSAS and the human graders. This metric was chosen since this was used in the official competition of ASAP-SAS. Other works \cite{chen2013automated,ramachandran2015identifying} also used it to evaluate their results. It calculates the level of agreement between the two raters. It also takes into account the \emph{by chance} probability of assigning the same grade to a sample by both the raters.  

Quadratic Weighted Kappa is calculated as follows. Firstly, the weight matrix \(W\) is constructed according
to Equation \ref{Weight_matrix}. Here \(i\) is the reference rating of human rater, \(j\) is the rating assigned by the model and \(N\) is the total number of possible ratings.
\begin{equation}
\label{Weight_matrix}
W_{i,j} = \frac{(i-j)^2}{(N-1)^2}
\end{equation}
After this, \textit{QWK} is calculated as:
\begin{equation}
\kappa = 1 - \frac{\Sigma_{i,j}W_{i,j}O_{i,j}}{\Sigma_{i,j}W_{i,j}E_{i,j}}
\end{equation}

Here matrix \(O\) contains the observed scores such that rating \(i\) is given by human grader and \(j\) is given by the model. $W_{i,j}$ contains the weights as derived in Equation \ref{Weight_matrix} and \(E\) contains the expected scores obtained by multiplying the histogram vectors of the two scores \emph{i.e.} the ones by human graders and the other by the proposed model AutoSAS. Subscripts in Matrix $O_{i,j}$ correspond to the number of essays that score \(i\) from the first rater and \(j\) from the second one. 

\subsection{Experimental Setup} We conducted the experiments on a machine with Intel(R) Core(TM) i5-3210M CPU @2.50GHz with a 10.0 GB of RAM and Operating System as 64-bit Windows 10. The Internet speed is close to 8 Mbps. On this system, it took slightly lesser than 25 minutes (average time) to extract features, train and test the model for a particular set of question prompt. This indicates that AutoSAS could be used by common students, schools, teachers and MOOCs alike without much overhaul of their existing systems. Such scoring can even happen on a teacher's or a student's personal computer and not just on a powerful laboratory computer.  

\subsection{Discussion} The results 
for the evaluation of AutoSAS and those of the models used by \cite{ramachandran2015identifying} and \cite{riordan2017investigating} are presented in Table 2.  Ramachandran \textit{et al.} used word order graph in order to capture the order of the tokens and lexico-semantic matching technique for identifying the degree of relatedness across tokens and phrases. They replaced the manually coded patterns used in the best performing model in Kaggle with the automatically generated patterns produced by their method, and used them as features for training a Random Forest model. Riordan \textit{et al.} used neural networks with n-grams and word embeddings as features. The performance of their systems are directly reported from their papers.  As shown, AutoSAS outperforms \cite{riordan2017investigating} on each of the prompts. AutoSAS also outperforms model given by \cite{ramachandran2015identifying} on 6 out of 10 sets. In the remaining 4 of the short answer sets, it performs equally well in 2 of them, in one of the sets it performs slightly worse and in set 6 it lags behind the other models.

AutoSAS performs exceptionally well on set 3, performing 8.8\% better than the current best model of \cite{riordan2017investigating}. The question for set 3 asks students to explain the similarities between Pandas in China and Koalas in Australia and how they are different from Pythons. This question demands some specific details from the students as its answer. These details can be found in the comprehension question prompt given to them. Thus majority of the answer can be derived from the question itself, but many responses go beyond the details mentioned in the question. They include some details which cannot be derived just from the question but require some prior knowledge. This is what most hand-tailored approaches and features such as those used by \cite{ramachandran2015identifying} fail to grasp. Prior knowledge of the subject can be fed into the training models only by introducing it to the texts that contain those specific concepts and facts. Only then it can effectively grasp what the student have written, and in the process, modeling what the teacher would have done when faced with a similar scenario. AutoSAS does this task by acquiring information that is outside the purview of the prompt using the features such as \textit{Weighted Keywords} and \textit{Word2Vec/Doc2Vec} embeddings.

Next, we show the feature importance and ablation study. It is worthy to note that neither of the works \cite{ramachandran2015identifying,riordan2017investigating} with which we compare our performance have conducted such a study. Thus an important aspect of the grading process is absent from the present state-of-the-art systems as reported to the research community.

\subsection{Ablation Study} Table 3. presents the findings of various feature groups. It lists the rankings of the feature groups as well as the fall in accuracy observed after removing the said features. For getting the fall in accuracy value for a particular group, the features extracted from that group were removed while keeping the other groups intact. The smaller set of features are fed to the Random Forest model and the results computed are presented. 

\textit{Word2Vec} and \textit{Doc2Vec} were the most important features. This might be due to the fact that the dataset is generic in nature and is represented well by these embeddings. Features that are based on in-domain information are also highly valued. The examples of such features are \textit{prompt information}, \textit{weighted keywords}, \textit{lemmatized response} and \textit{lexical overlap}. 

The additional features such as \textit{word frequency}, \textit{difficulty}, \textit{statistics of word} and \textit{sentence length} do not figure highly neither in the rankings nor are the accuracy values being affected significantly. But, in any case, they do prove to be useful for predicting the scores. 

Although, it was mentioned that the graders did not consider word frequency, penmanship, for grading a particular response, but as indicated, these biases do show up in the gradings either knowingly or unknowingly. With AutoSAS, it is a virtue of the system that one can turn off these features if one does not want to take into consideration these specific details.

\begin{table}[htbp]
\label{comparison_table_2}
\begin{tabular}{|l|l|c|}
\hline
\multicolumn{1}{|c|}{\textbf{Rank}} & \multicolumn{1}{c|}{\textbf{Feature Group}} & \textbf{\begin{tabular}[c]{@{}c@{}}Fall in \\ Accuracy\end{tabular}} \\ \hline
\begin{tabular}[c]{@{}l@{}}1\\ 2\\ 3\\ 4\\ 5\\ 6\\ 7\\ 8\\ 9\end{tabular} & \begin{tabular}[c]{@{}l@{}}Word2Vec, Doc2Vec\\ Prompt Overlap\\ Weighted Keywords\\ POS Tags\\ Lexical Overlap\\ Logical Operators\\ Temporal Features\\ Stats of Sentence and Word Length\\ Word Freq, Difficulty\end{tabular} & \begin{tabular}[c]{@{}c@{}}23.54\%\\ 20.85\%\\ 16.93\%\\ 12.36\%\\ 8.45\%\\ 6.40\%\\ 4.2\%\\ 2.11\%\\ 1.02\%\end{tabular} \\ \hline
\end{tabular}
\caption{Importance of various features used in AutoSAS.}
\end{table}

\section{Conclusion and Future Work} In this work, we proposed a supervised regression model and explored different linguistic features for grading short answers, and a system encompassing it named AutoSAS, which can be easily used and deployed in various educational and professional testing settings. Experiments on the publicly available dataset ASAP-SAS showed that AutoSAS outperforms the current state-of-the-art algorithms and approaches. According to \cite{powers2000comparing}, the agreement between machine learning models and expert human graders range between 0.7 to 0.8, and AutoSAS achieved a mean QWK score of 0.79. 

We also showed how AutoSAS can be useful for assessing the decision making process for assignment of a score and provide valuable feedback to the users about the characteristics of a response. Unlike the existing state-of-the-art systems, we perform an ablation study and discuss about the most important features that contribute towards the performance of our trained model. One of the major aspects where AutoSAS still lacks is review comments. We would like to work on it in the future and also try out hybrid methods that takes into account the Random Forest model along with a deep neural network architecture in order to improve our current system.

\bibliographystyle{aaai}
\bibliography{sample-bibliography}

\end{document}